\documentclass[10pt,twocolumn,letterpaper]{article}

\usepackage{iccv}
\usepackage{times}
\usepackage{epsfig}
\usepackage{graphicx}
\usepackage{amsmath}
\usepackage{amssymb}
\usepackage{booktabs}
\usepackage{makecell}
\usepackage{xcolor}
\usepackage{pifont}
\usepackage{caption}

\usepackage[pagebackref=true,breaklinks=true,letterpaper=true,colorlinks,bookmarks=false]{hyperref}

\iccvfinalcopy 

\def\iccvPaperID{5254} 

\ificcvfinal\fi

\begin{document}

\title{GlobalMapper: Arbitrary-Shaped Urban Layout Generation}

\author{Liu He \\
Purdue University\\
{\tt\small he425@purdue.edu}
\and
Daniel Aliaga\\
Purdue University\\
{\tt\small aliaga@purdue.edu}
}



\twocolumn[{%
\renewcommand\twocolumn[1][]{#1}%
\maketitle

\setlength{\tabcolsep}{1.0pt}
\centering
\begin{tabular}{ccccc}
    \captionsetup{type=figure}
    \includegraphics[width=0.24\textwidth]{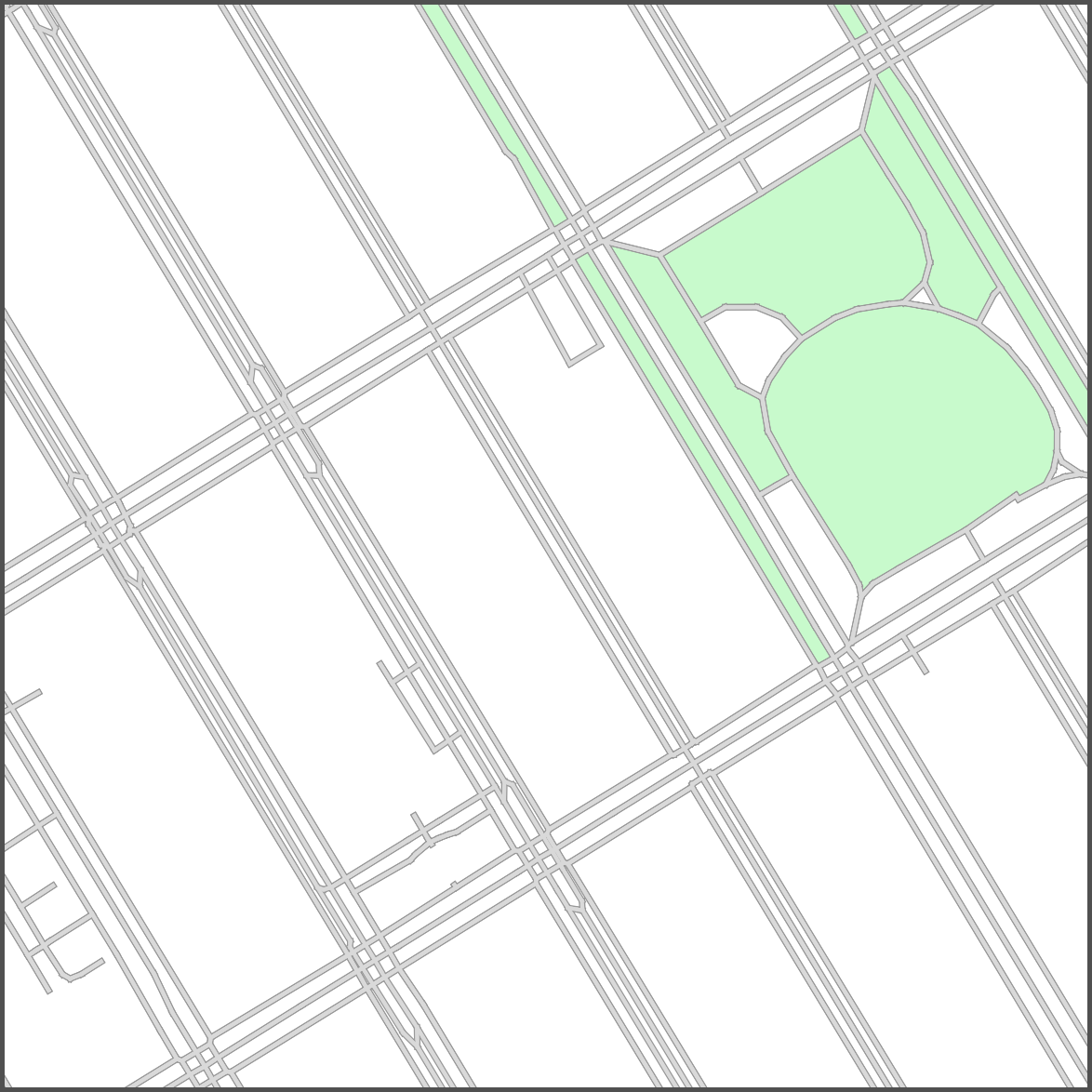} &
    \includegraphics[width=0.24\textwidth]{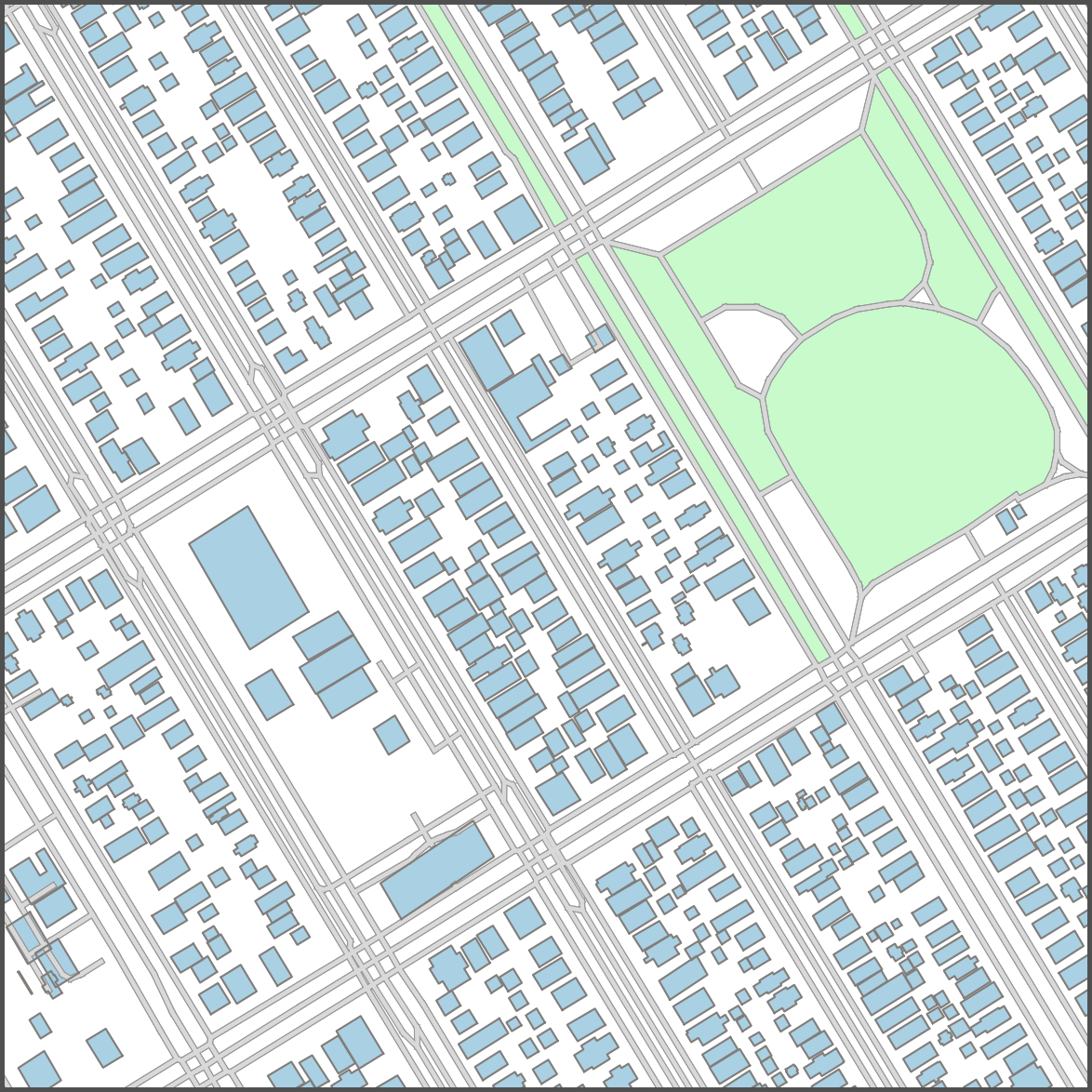} &
    \quad \quad &
    \includegraphics[width=0.24\textwidth]{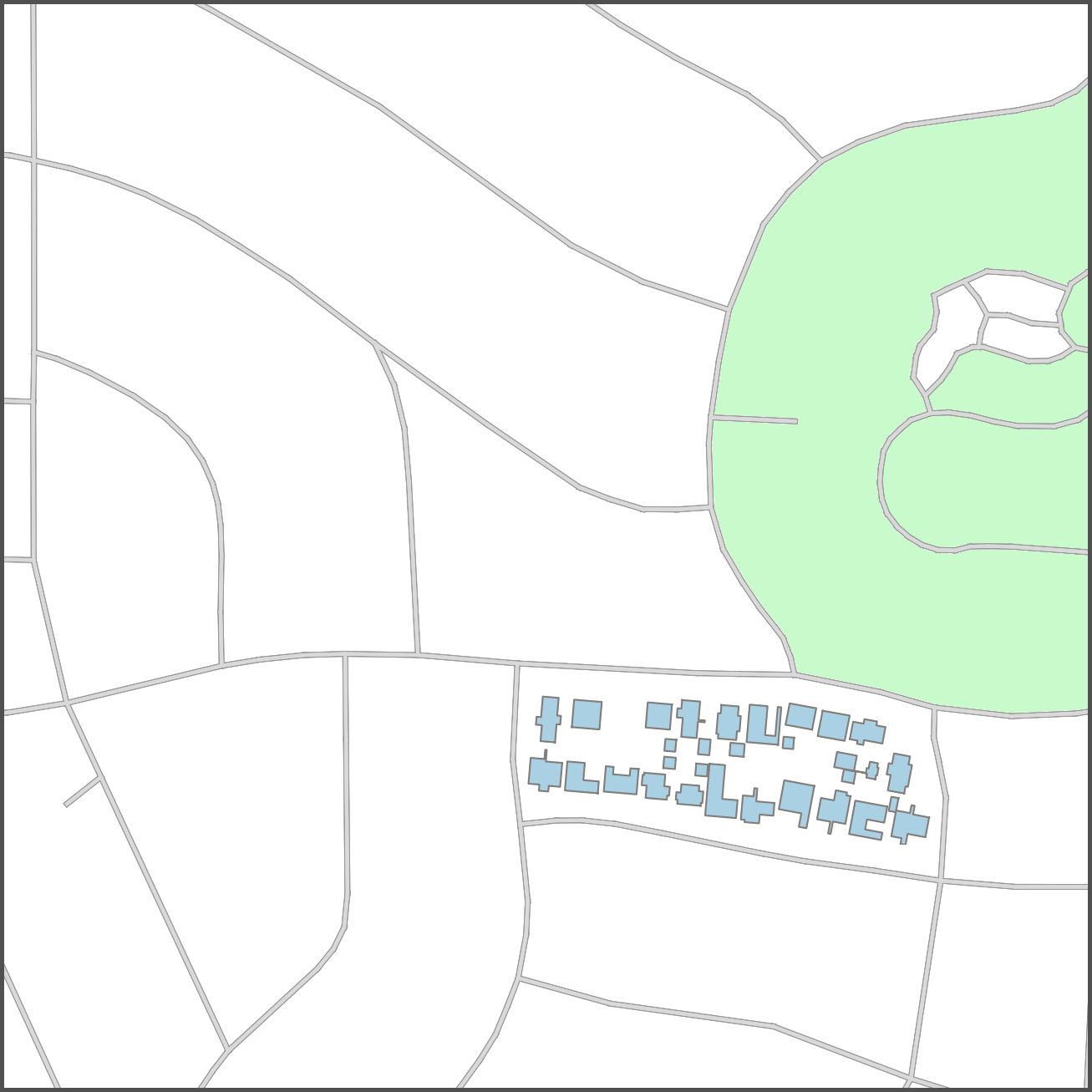}  &
    \includegraphics[width=0.24\textwidth]{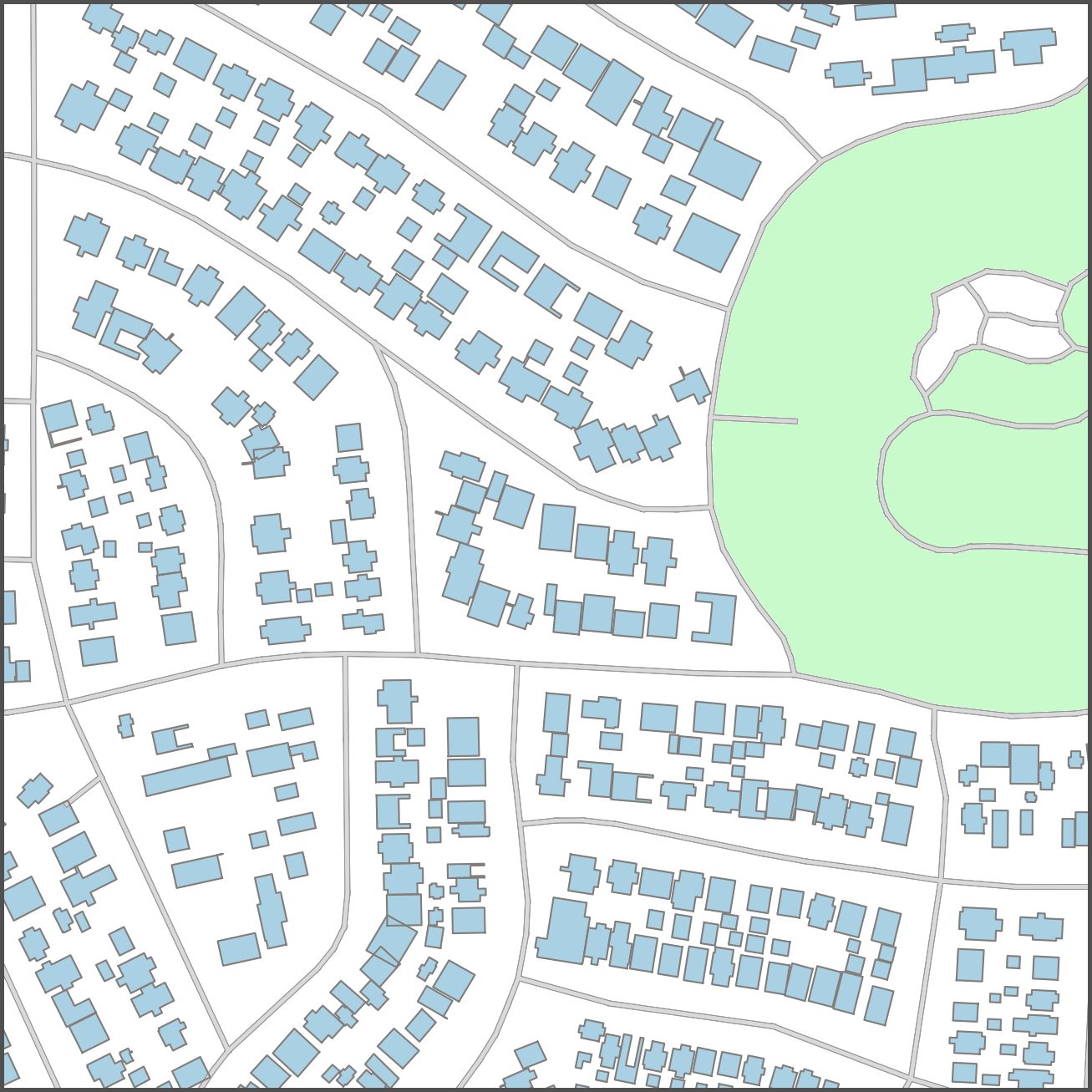}      
    \\

 Road Network & Random Generation & \quad & Road Network + Prior & Conditional Generation
\end{tabular}%
\captionof{figure}{\textbf{Arbitrary Urban Layout Generation.} Our method generates realistic urban layouts given an arbitrary road network (Left Pair). Moreover, given learned prior from a building layout, our method conditionally generates similar urban layouts for the given arbitrary road network (Right Pair). } \label{fig:teaser}
\bigbreak
}]

\ificcvfinal\thispagestyle{empty}\fi

\begin{abstract}
Modeling and designing urban building layouts is of significant interest in computer vision, computer graphics, and urban applications. A building layout consists of a set of buildings in city blocks defined by a network of roads. We observe that building layouts are discrete structures, consisting of multiple rows of buildings of various shapes, and are amenable to skeletonization for mapping arbitrary city block shapes to a canonical form. Hence, we propose a fully automatic approach to building layout generation using graph attention networks. Our method generates realistic urban layouts given arbitrary road networks, and enables conditional generation based on learned priors. Our results, including user study, demonstrate superior performance as compared to prior layout generation networks, support arbitrary city block and varying building shapes as demonstrated by generating layouts for 28 large cities.

\end{abstract}
\section{Introduction}

\begin{figure*}[h!]
  \includegraphics[width=\linewidth]{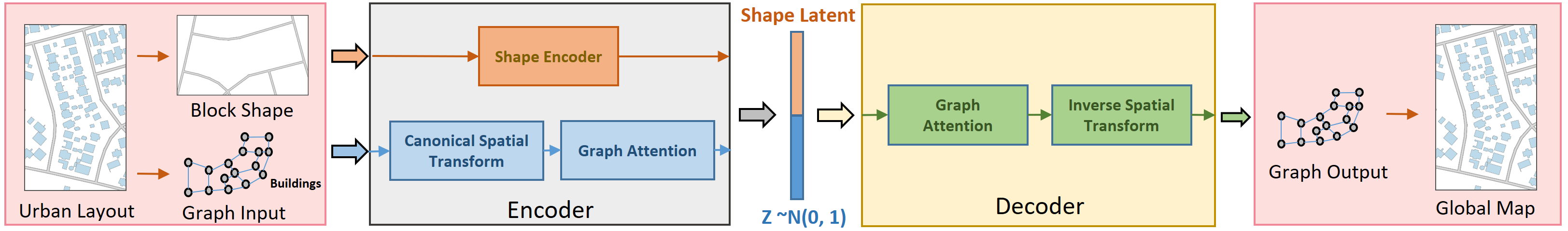}
  \caption{\textbf{Framework.} Our method is trained in a VAE framework conditioned on arbitrary block shapes (e.g., from a road network). The block shape is a binary mask encoded to latent space by a CNN-based encoder. Building layouts are represented by a grid topology graph and its positional, geometrical, and shape information is encoded by a spatial transform into a canonical representation. By using a graph attention network we codify the graph into a latent space. During decoding, the block shape latent is utilized as condition for decoder to produce building layouts. Our method is able to conditionally generate large urban layouts for arbitrary block shapes and building arrangements (i.e., we have processed 28 cities).}
  \vspace{-1mm}
  \label{fig:pipeline}
\end{figure*}

Layout generation is of significant interest today in computer vision, computer graphics, and related fields. In particular, an urban building layout consists of a set of buildings arranged into arbitrarily shaped city blocks as defined by a network of interconnected roads. Such layouts are needed in order to provide urban configurations for entertainment, simulation, and scientific applications such as designing and evolving cities to address urban weather forecasting (Fig.~\ref{fig:sparse-prior}), urban heat island modeling, and sky view factor analysis, etc.



The capture and modeling of complex urban building layouts has been pursued from several directions. 1) Computer vision and photogrammetry works seek to metrically reconstruct building layouts over large areas from satellite and/or aerial imagery~\cite{cheng2019darnet, musialski13, demir2018deepglobe, he2023generative, xu2022cvnet, zorzi2022polyworld, li20213d, mahmud2020boundary, firoze2023tree}. Nonetheless, these approaches are constrained by image resolution, occlusion, and segmentation accuracy. 2) Urban procedural modeling has made significant strides in generating the hierarchical components of a city such as road networks, city blocks, parcels, buildings, and more (e.g., buildings~\cite{muller2006procedural,zhang2022guided,bhatt2020design}, roads~\cite{nishida16, chen2008interactive}, trees~\cite{he2016automatic,huang2020part,li2021learning,zhou2023deeptree}, parcels~\cite{Lipp11, vanegas2012procedural} and even entire cities~\cite{vanegas12, benes2021urban}). Unfortunately, the generation process requires time consuming and expert knowledge in developing custom procedural rules to create or update the content. 3) Inverse urban procedural modeling attempts to remove the need of explicit rule specification by encoding heuristics specific to cities~\cite{bokeloh10, zhang20}.





Recently, some solutions have addressed inverse modeling for layout generation; e.g., using graph-based methods for structured object creation (StructureNet~\cite{mo19}), interior layout design (HouseGAN~\cite{Nauata20}), land lot generation (BlockPlanner~\cite{xu2021}), and building generation (Building-GAN~\cite{chang2021building}). However, none of these works focus on generating urban layouts containing arbitrary city block shapes and varying building footprints such as those appearing in cities worldwide.






Our work builds on three key observations. 
\begin{itemize}
    \item First, following the organization defined by procedural modeling arbitrarily-shaped buildings within an arbitrarily-shaped city block can be organized as a discrete spatial data structure of non-overlapping objects arranged into a few rows of buildings (e.g., one row for large buildings, two rows for dense smaller structures, and multiple rows for sparser areas having main buildings and auxiliary building structures).
    


    \item Second, capturing the aforementioned building layout configuration can be performed with at least a 2D message passing setup as afforded by a stubby grid graph (i.e., one row of connected nodes/buildings, two rows forming a ladder graph, or a few rows interconnected as a shallow regular grid, as in Fig.~\ref{fig:spatial transform}. 


    \item Third, by using skeletonization we can perform a spatial transform to the aforementioned stubby graph topology to support diverse urban block shapes. 
\end{itemize}

Altogether, our fully automatic approach for building layout generation uses a variational auto-encoder framework using graph neural networks. First, we define a graph-based representation of a block and building layouts that is able to capture the inherent styles of multiple cities globally. Second, the encoder makes use of multi-pass aggregation and combination message passing as well as a spatial transform to codify varying building shapes and layouts for arbitrary city block shapes (Fig.~\ref{fig:teaser}). Third, the decoder generates varying building shapes and layouts from the latent space conditioned on block shape (i.e., the distribution of building layouts depends on the block shape). Unlike most prior graph-based method, our scheme is able to accurately capture block shape, building shape type, positional data, and alignment as well. Finally, our synthesis process generates arbitrary maps based on the generated graph, potentially spanning a large continuous area (e.g. a entire city).

To the best of our knowledge, our method is the first to generate building layouts in arbitrary city blocks, and is able to span diverse styles of 28 large cities across North America (e.g., Chicago, Los Angeles, New York City... full list in Sec.~\ref{sec:Training}). In Fig.~\ref{fig:teaser}, our method generates realistic urban layouts given arbitrary shaped road networks, and it provides controllable generation given building layout priors. We provide comparisons to several prior related methods~\cite{arroyo2021variational,gupta2021layouttransformer,Jyothi19,xu2021,inoue2023layoutdm,he2023diffusion} and show that none of them supports arbitrary block shapes, various building shapes and layouts, nor have any been demonstrated at the scale of our method. In addition, we show examples of conditioned generation, manipulation, interpolation, and ablation study.

Our main contributions include:
\begin{itemize}
\item \textit{Graph-based Representation} for arbitrarily shaped city blocks and building layouts amenable to deep learning.
\item \textit{Canonical Spatial Transformation} to encode building layouts into a shape-independent canonical representation,
\item \textit{Controllable Generation} of realistic urban layouts for arbitrary road networks (e.g. random, or conditioned on a learned prior).
\end{itemize}



\section{Related Works}

The task of urban layout analysis from images as well as design and modeling has a long history. The seminal paper of~\cite{parish01} fomented urban procedural modeling, including that of building masses~\cite{muller2006procedural}, urban layouts~\cite{aliaga2008interactive, groenewegen2009procedural, Lipp11}, and parcel generation (e.g., \cite{vanegas2012procedural}, Esri CityEngine). This last work, as well as~\cite{carmona2021public}, describe that the building layouts of a city block usually consist of one of two fundamental styles, both of which are afforded by our approach but without the need to manually provide the procedural template. In addition, interior layout approaches have also been developed such as for furniture layout~\cite{Yu11}, mid-scale interior layouts~\cite{Feng16} and building interiors~\cite{merrell10}.


Recently, with the rise of deep learning~\cite{wang2018msnet,sheng2021ssn, sheng2022controllable, song2023objectstitch, sheng2023pixht}, numerous other data-driven and network-based layout generation approaches have been proposed, yielding improved performance and automation once trained. LayoutGAN~\cite{Li21} and LayoutVAE~\cite{Jyothi19} present general 2D layout planning tools. Alternative works~\cite{patil2020read, tabata2019automatic, kikuchi2021constrained} focus on document or graphic layout generation. Recent advancements in self-attention networks (e.g. Transformer) have been applied to natural scene and document layout generation by~\cite{gupta2021layouttransformer, arroyo2021variational, yang2021layouttransformer}. Latest diffusion models are also introduced to document layout generation~\cite{inoue2023layoutdm,he2023diffusion}. Other methods have focused on urban scenarios. For example, Building-GAN~\cite{chang2021building} generates 3D building structures, HouseGAN~\cite{Nauata20} arranges rooms within a single floor of a house. Works by Ritchie \etal~\cite{Ritchie19} and Wang \etal~\cite{Wang18, Wang19} synthesize the layout of indoor scenes. Further,~\cite{Wu19, para2021generative} generate interior plans for residential buildings. However, all previous works assume layout generated on a rectangular canvas. Moreover, the position and geometry of the predicted bounding box is scaled to a finite set of categories (e.g. image coordinates~\cite{Jyothi19}, vocabulary table~\cite{gupta2021layouttransformer, arroyo2021variational,inoue2023layoutdm,he2023diffusion}). It benefits from rectilinear alignment in some tasks (e.g. document layout generation), but constrains the randomness of building geometry as it may appear anywhere in a city block. To the best of our knowledge, our method is the first to handle arbitrary block-shapes/canvas and to include varying building/bounding-box shapes.


The approaches most relevant to our work are the building-layout generator of~\cite{Bao13} and the BlockPlanner method of~\cite{xu2021}. Bao \etal~\cite{Bao13} assumes that a set of hard and soft constraints are provided as input, and then generates a single building layout and proposes a methodology to evaluate the ''goodness'' of a layout with respect to the provided constraints. While our goal is also to generate building layouts, we seek a multi-building layout spanning an entire city block based on a conditioned input or a provided style, and then perform this task at scale (e.g., a fragment or more of a city). Further, we wish the style, which is analogous to the user-provided constraints in~\cite{Bao13}, to be inferred from example data and not provided manually. 

BlockPlanner~\cite{xu2021} generates only rectangular building shapes inside rectangular city blocks. Their graph-based solution generates an approximate solution and then uses a refinement step to obtain zero-gap building output, which is an assumption of theirs (i.e., buildings in dense parts of cities). In comparison, our method adds canonical spatial transform to support arbitrary block shapes, can represent various building shapes, uses a grid topology to afford more complex building layouts rather than only a ring topology, employs attention-based networks instead of strict message passing, supports more buildings per block, and is demonstrated on 28 large cities instead of only 1 (New York City).

Nonetheless, in Sec.~\ref{sec:comparison} we perform explicit comparisons to the domain-specific method BlockPlanner~\cite{xu2021}, to the natural scene layout method LayoutVAE~\cite{Jyothi19}, to self-attention networks Gupta \etal~\cite{gupta2021layouttransformer} and its variational modification~\cite{arroyo2021variational}.
 In all cases, our approach outperforms these existing methods qualitatively and quantitatively.

\section{Building Layout Generation}

We describe our approach to deep building layout generation (Fig.~\ref{fig:pipeline}). First, we summarize our graph representation and normalization process for arbitrary city blocks. Second, we provide details on our encoding phase. Third, we describe our conditional decoding/generation phase. Finally, we provide information on our synthesis phase as well as auxiliary algorithms.

\subsection{Graph Representation}
\label{sec:Representation}

\begin{figure}[t]
  \includegraphics[width=\linewidth]{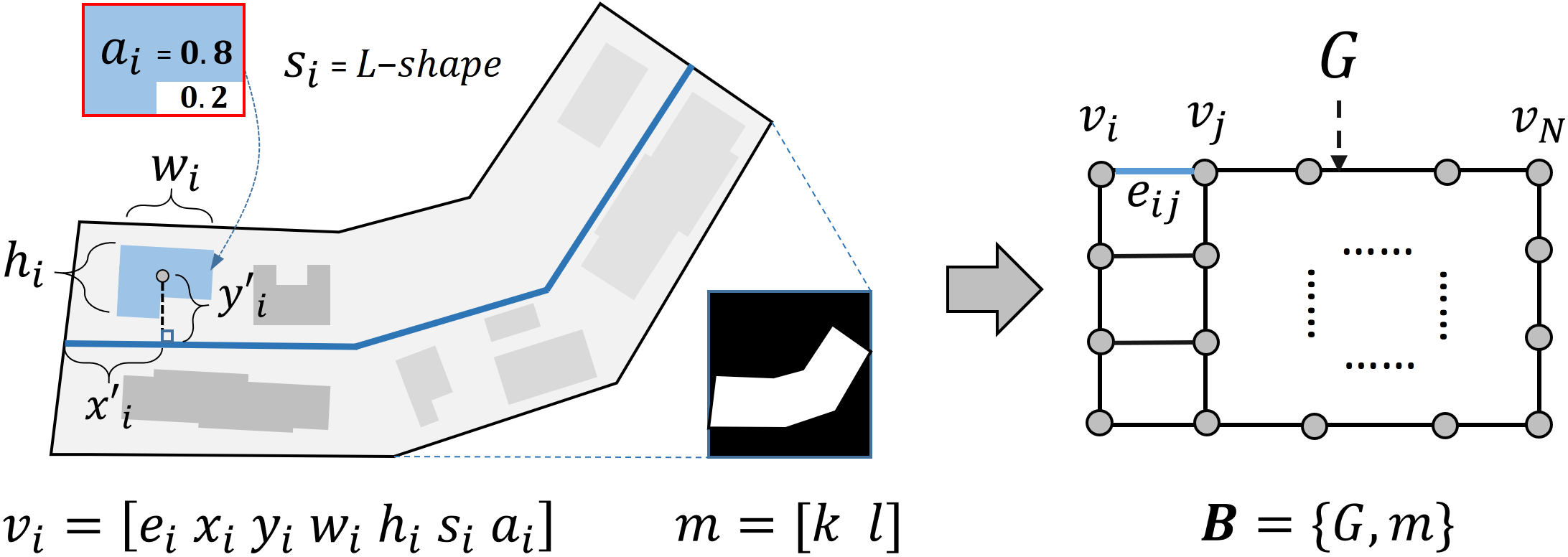}
  \caption{\textbf{Graph Representation.} Our method represents any city block $B$ using a normalized stubby graph representation $G$ (i.e., a grid graph with many columns and few rows), 
  and a set of graph-level features $m$.
  \textbf{Canonical spatial transformation} finds the main axis (blue line) of an arbitrary city block, and transforms each node's spatial information by the relative distance to the main axis.}
    \vspace{-1mm}
  \label{fig:spatial transform}
\end{figure}

A urban layout is defined by a network of roads that forms city blocks and within each city block a set of buildings. In Fig.~\ref{fig:spatial transform}, each city block $B$ is represented by a graph $G$ of the multiple buildings within the block and by a set of block shape features $m$.

\textbf{Buildings.} We represent the building layout of an arbitrary city block $B$ using a stubby grid graph $G$ (e.g., the graph has many more columns than rows and each vertex typically has 2 or 3 incident edges). We define the graph as $G = \{ V, E \}$,  where $V$ is a set of node vertices $V = \{v_i\}$, $i\in [1, N]$ that each vertex $v_{i}$ describes each building, and $E$ is a set of edges $E = \{ e_{ij}\}$, $i,j \in [1, N]$ between spatially adjacent nodes/buildings. $N$ is defined as the maximum building count in one block. We empirically set $N = 120$ to handle $>$99\% percent of real city blocks in our dataset.

Each vertex $v_{i}$ stores a building's geometric and semantic feature information. In particular, the per-vertex (or building) features are the following:
\begin{itemize}
    \item \textbf{$e_i$} is the binary existence flag of this building (e.g., a value of one implies the building exists, else zero).
    \item \textbf{$(x_i,y_i)$} is the physical position of building center.
    \item \textbf{$(w_i, h_i)$} is the width and height of the building's oriented 2D bounding box (the width dimension is parallel to the main axis),
    \item \textbf{$s_i$} is an integer representing the building shape type (described below), and
    \item \textbf{$a_i$} is the occupancy ratio of the building area to the oriented 2D bounding box area.
\end{itemize}

The aforementioned features $\{s_i, a_i\}$ represent each building by a shape type and an occupancy ratio. In a preliminary analysis, we found that buildings in our test areas can be fit with IoU=$95.78\%$ and Hausdorff Distance $=1.36m$ (i.e., the average of maximum distance between the contour of the actual building and our best-fit polygonal building shape is $1.36m.$) using one of four fundamental parameterized shape types: \textit{Rectanglar, L-shape, U-shape and X-shape}. During the preprocessing phase, we use Powell optimization to determine the best-fitting building shape type and calculate an occupancy ratio (see Supplemental for more details). Both variables enable our later-described synthesis step (Sec.~\ref{sec:Synthesis}) to generate a synthetic building footprint.



\textbf{Blocks.} The block shape features $m$ are composed of a binary mask $k$ and the mask scale $l$.  We typically use a mask resolution of 64x64 pixels. The longest side of each block's oriented bounding box is rotated to horizontal and scaled to fit within the mask (see Fig.~\ref{fig:spatial transform}).

\subsection{Variational Learning}
\label{sec:Encoding}


In this section, we first describe the spatial transform of an arbitrary urban layout into a canonical form, and then describe the shape encoder setup which enables using block shapes as a condition of the decoder setup (Fig.~\ref{fig:pipeline}). We also discuss our variational model structure and training scheme. 

\textbf{Canonical Spatial Transform}. In order to support a wide variety of urban layouts at arbitrary scales, we introduce a transformation parameterized by block shape feature $m$. It maps each building graph $G$ to a canonical representation $G'$. Prior urban procedural modeling work~\cite{vanegas2010building} indicates that urban layouts follow typical conventions which we exploit to define the canonical representation. In particular, using the contour of block mask $k$ we compute a block's 2D skeleton and modify its endpoints to obtain an extended skeleton as a compact polyline structure forming a main axis (Blue line in Fig.~\ref{fig:spatial transform}). Then, for each building in $G$ we compute its normalized position ($x'_{i}, y'_{i}$ in Fig.~\ref{fig:spatial transform}) relative to one of the main axis polyline segments, and to one of the minor axis perpendicular to a main axis line segment. The length of main axis and minor axis is utilized to normalize building's position and size, ultimately producing the canonical $G'$. This spatial transform provides a regularized spatial representation of building layout regardless of block shape as compared to direct scaling by most former relative methods~\cite{xu2021, Jyothi19, gupta2021layouttransformer, arroyo2021variational}. Additional details are in Supplemental.

\textbf{Shape Encoder.} To obtain a latent representation of block shape feature $m$, a CNN-based autoencoder is trained to reconstruct the binary mask $k$. The block scale $l$ is stacked with the binary mask as the second channel of the input image. The encoder model has 4 layers of convolutional layers and the corresponding decoder model is the mirror image of this. The model is able to reconstruct a binary mask with IoU $>$ 0.98. During our variational training, the latent representation $m'$ encoded by the shape encoder can be utilized as a conditional prior of the decoder.



\textbf{Conditional Variational Training.}
As a conditional VAE, our model aims to represent the conditional data distribution of $p(G'|m')$, which is intractable to compute. By variational learning, the model approximates the distribution by maximizing its evidence lower bound (ELBO). The objective function is as follows:

\begin{equation}
\footnotesize
\label{eqn:objective}
\mathcal{L}(\theta, \phi) = \underset{q_{\phi}(z|G', m')}{\mathbb{E}}[\log p_{\theta}(G'|z,m')] - \mathrm{KL}[q_{\phi}(z, m'|G') \| p(z)]
\end{equation}

where the approximate posterior $q_{\phi}(z|G', m')$ is parameterized by $\phi$, and the decoder $p_{\theta}(G'|z,m')$ is a deep neural network parameterized by $\theta$. The prior distribution $p(z)$ is a standard Gaussian distribution in our training. Particularly, our model should be able to encode and reconstruct $G'$ conditioned by $m'$.

\textbf{Graph Attention.} We use a graph attention network~\cite{gat2018} (GAT) as the backbone of our encoder and decoder. Graph attention networks perform weighted multi-layer messaging passing between connected nodes. In particular, the edges of our 2D grid graph topology enable message passing between buildings next-to and in-back-of other buildings. We denote node features in the canonical graph as $f_i$. Our formulation can be represented by:


\begin{equation}
\label{eqn:convolution}
f_i^t = GAT(f_i^{t-1})
\end{equation}

We collect all individual node feature vectors into feature matrix $F^t = \{f_i^t\}, i \in [1, N], t\in[1,T]$ and apply equation Eq.~\ref{eqn:convolution} a total of $T$ times. Then, the feature matrices from all $T$ iterations are aggregated as $[F^t, F^1, ...., F^T]$ to form a 512 dimensional normal distribution that is sampled by the $z$ of a variational reparameterization. 


The decoder $p_{\theta}(G'|z,m')$ is conditioned by $m'$. As Fig.~\ref{fig:pipeline} indicates, we concatenate $m'$ with $z$ as the input of our decoder. Initially, a sampled latent space vector $z$ is pushed through a MLP to produce an initial feature matrix $\hat{F^0} = \{\hat{f}_i^{0}\}, i \in [1, N]$, which consists of single feature vector for each node vertex/building. Then, a similar multi-layer messaging passing, using Eq.~\ref{eqn:convolution}, is performed $T$ times yielding $\hat{F}^T$. Finally, each vector component $\hat{f}_i^T$ within $\hat{F}^T$ is decoded by independent MLPs for predictions of each building feature element in $\hat{G'}$ (e.g., $\{e_i, x_i, y_i, w_i, h_i, s_i, a_i\}$). Predicted features are calculated by reconstruction loss. The weighted sum of each loss component is back-propagated to graph attention networks for weight updating.

\textbf{Inverse Spatial Transform.} With block shape feature $m$, the spatial information of $\hat{G'}$ (e.g., building position, width, and height) is inversely transformed back to its original scale and location.

\subsection{Synthesis}
\label{sec:Synthesis}

In this phase, we generate building shapes within each generated city block. Our goal is to produce building footprints that are similar (but not necessarily identical) to the original building coverages, positions, and shapes. Further as shown in Sec.~\ref{sec:Control-Map}, we can produce such an output using only a small fraction of the real-world data. 

Our method synthesizes the aforementioned four building shape types (\textit{Rectangular, L-shape, U-shape, X-shape}) using a parameterized function (see Supplemental for details). For the buildings of a city block, the shape parameters of each generated building type $s_i$ are randomly perturbed within a predefined range until a configuration best satisfying the generated occupancy value $a_i$ is found within a maximum number of iterations. Then, the generated building footprint is rotated  so that its width dimension is again parallel to the block's main axis.
 



\section{Experiments}
\label{sec:experiemnt}
\subsection{Training}
\label{sec:Training}
\textbf{Datasets.} We collected datasets from 28 large cities across North America (Chicago, D.C., New York City, Atlanta, Dallas, Los Angeles, Miami, Seattle, Boston, Providence, Baltimore, San Diego, San Francisco, Portland, Milwaukee, Minneapolis, Austin, New Orleans, Phoenix, Denver, Toronto, Vancouver, Pittsburgh, Tampa, San Jose, Norfolk, Austin, and Houston). The total number of city blocks is 119,236, containing 2,513,697 buildings, which is 187-times bigger than the training set of~\cite{xu2021}. We split the data 80\% for training, and 20\% for validation. We obtained the datasets by downloading 2D layouts from OpenStreetMap (OSM)~\cite{OpenStreetMap}. Minimal clean-up was done to simplify parallel multi-lane roads to single edges and to remove self-looping edges.

\textbf{Training Details.} We performed an end-to-end training of our pipeline.
After various experiments, we set learning rate $lr = 0.001$, and each training typically converges in $10^5$ iterations (9 hours on a single NVIDIA A5000 GPU). 
We determined the best configuration of loss functions to be using $L2$ loss for all geometry features $\{x_i, y_i, w_i, h_i\}$, and cross entropy loss for categorized features $\{e_i, s_i\}$, and KL-divergence loss for $z$. The relative loss weight for geometry features, categorized features, and KL-divergence loss is 4.0, 1.0, 0.5, respectively. We found additional regularization losses (i.e., a loss penalizing overlaps~\cite{xu2021}) do not further reduce errors. The average inference time is 5.81ms per city block by a single A5000 GPU.


\begin{table*}[t]
\small
\centering
    \vspace{-1mm}
\begin{tabular}{lccccccc}
    \toprule
   Method & L-Sim$\uparrow$  & Overlap$\downarrow(\%)$ & Out-Block$\downarrow(\%)$ & FID$\downarrow$ & WD$\downarrow(bbx)$ & WD$\downarrow(count)$  \\
    \midrule
   LayoutVAE~\cite{Jyothi19}    & 4.49 & 33.39 & 11.15 & 94.54 & 7.24 & - \\
   BlockPlanner~\cite{xu2021} & 14.92 & 9.46 & \underline{2.24} & \underline{39.27} & 6.20 & \textbf{0.03} \\
   Gupta \etal~\cite{gupta2021layouttransformer} & \underline{17.59} & 3.61 & 7.58 & 47.06 & \underline{2.23} & 6.12  \\
   VTN~\cite{arroyo2021variational} & 17.65 & \underline{1.49} & 7.97 & 46.71 & 2.78 & 3.98 \\
    \midrule
   \textbf{Ours} & \textbf{22.45} & \textbf{1.42} & \textbf{0.89} & \textbf{14.94} &  \textbf{1.45}  & \underline{0.06}\\

    \bottomrule

\end{tabular}
    \vspace{-2mm}
  \caption{\textbf{Quantitative Results.} We generate 1000 urban layouts and compare to the same amount of real urban layouts. Best values are in bold, second best values are underlined. Our method outperforms other existing methods in 5 metrics.}
    \vspace{-2mm}
\label{tab:comparison}
\end{table*}%




\subsection{Comparisons}
\label{sec:comparison}
We compared our approach to 4 related layout generation methods: LayoutVAE~\cite{Jyothi19}, Gupta \etal~\cite{gupta2021layouttransformer}, VTN~\cite{arroyo2021variational}, and BlockPlanner~\cite{xu2021}. For first two methods, we utilized the codes provided by the author's repository of ~\cite{gupta2021layouttransformer}. We provided the correct bounding box count for LayoutVAE and only retrained its BBoxVAE portion to predict bounding box geometry. For~\cite{gupta2021layouttransformer}, we transferred our dataset to COCO format and retrained the network using default settings in repository. During generation, we setup their top-k sampling as k=5 to produce diverse layout outputs. For VTN~\cite{arroyo2021variational}, we modified the codes of ~\cite{gupta2021layouttransformer} by adding a variational training framework. For BlockPlanner~\cite{xu2021}, we only received partial model structure source code from the authors. We re-implemented the rest of the method to the best of our ability, and retrained it on our dataset by the suggested settings in their paper. Additionally, comparisons to diffusion models~\cite{inoue2023layoutdm,he2023diffusion} are provided in Supplemental.

\textbf{Quantitative Results.} We generated 1000 urban layouts by ours and by each existing method, and compared to the same amount of real urban layouts. Our method is conditioned on the road networks from real urban layouts. The LayoutVAE is conditioned on the building count we provided. VTN is conditioned on the corresponding real building layouts. Gupta \etal~\cite{gupta2021layouttransformer} generation is fully random. The set-to-set comparison is evaluated by 6 quantitative metrics. Overlap index~\cite{li2019layoutgan} is the percentage of total overlapping area among generated building layouts within the urban block. The Out-Block index is the percentage of generated building layout area that is out of the urban block contour. Wasserstein distance (WD) indicates the similarity between two distributions. Specifically, we computed the WD between the distributions of building counts and bounding box geometry (location and size, centered around the origin). Lower score indicates generated urban layouts provide better similarity to the real world distribution of building counts and geometry arrangement. We also computed FID score~\cite{heusel2017gans} to evaluate diversity and quality of the generated urban layouts.

Inspired by DocSim index~\cite{patil2020read}, we design a similarity metric LayoutSim (L-Sim) to evaluate the geometry similarity between pairs of urban layouts. We first rotate each of the two urban blocks to make the long side of its oriented bounding box horizontal, and translate to its center location without changing scale. Then we assign a matching score for all pairs of buildings from two different urban layouts. Pairs of buildings that are approximately overlapping and with similar sizes will have higher scores. Then we compute the maximum weight matching among all possible building pairs by Hungarian method~\cite{kuhn1955hungarian}. The average matching score is the L-Sim between two urban layouts. The formula for matching score between building $b_{1}$ and $b_{2}$ is as followed:

\vspace{-0.2in}
\begin{equation}
\label{eqn:LSim}
S(b_{1}, b_{2}) = MinArea(b_{1},b_{2}) \cdot 2^{-c\|b_1-b_2\|_{2}}
\end{equation}

The multiplier $MinArea(b_1, b_2)$ returns the minimum area of two buildings.  Exponent factor $\|b_1-b_2\|_{2}$ is the $L2$ norm between the positions of two buildings. Matching large buildings will tend to increase the matching score. Large position difference will decrease the value of the matching score. We set $c=0.02$ to scale the relatively large position difference value in the real world. 


As Tab.~\ref{tab:comparison} shows, our method clearly outperforms existing methods in 5 metrics and achieves the second best in WD of building counts. Note that we provide building count (the label set as described in original paper~\cite{Jyothi19}) to LayoutVAE. So it is unfair to us regarding the WD distance on building count. We don't report the value in the table.

\begin{figure*}
\centering
\setlength{\tabcolsep}{3pt}
\begin{tabular}{ccccc}
    \includegraphics[width=0.19\textwidth]{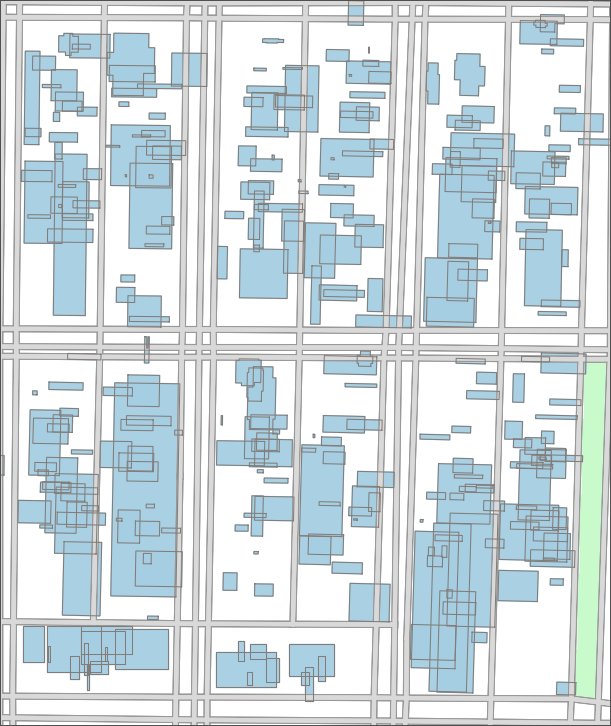} &
    \includegraphics[width=0.19\textwidth]{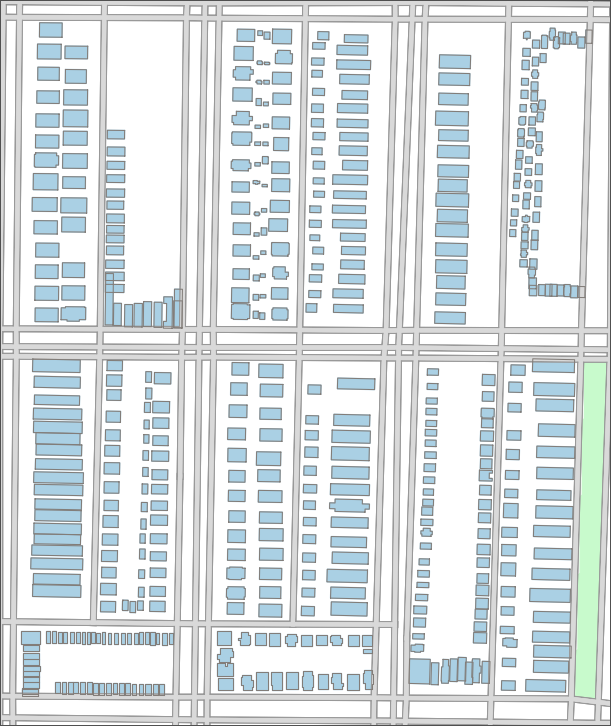} &
    \includegraphics[width=0.19\textwidth]{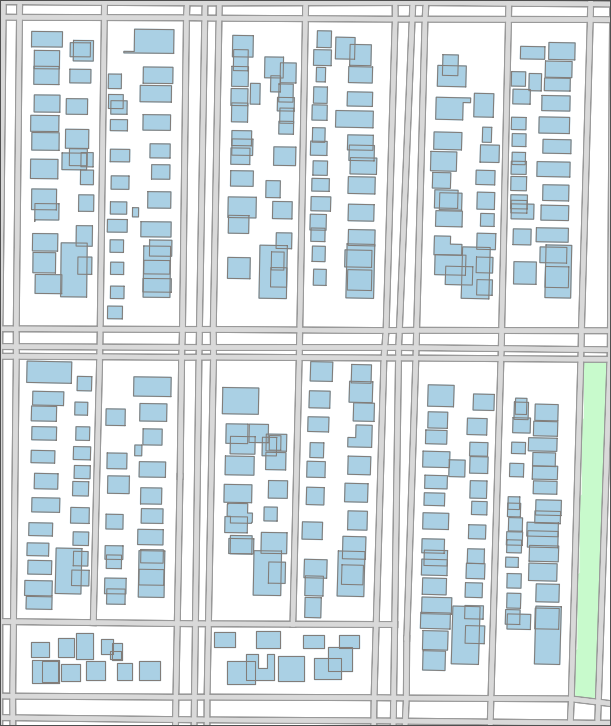} &
    \includegraphics[width=0.19\textwidth]{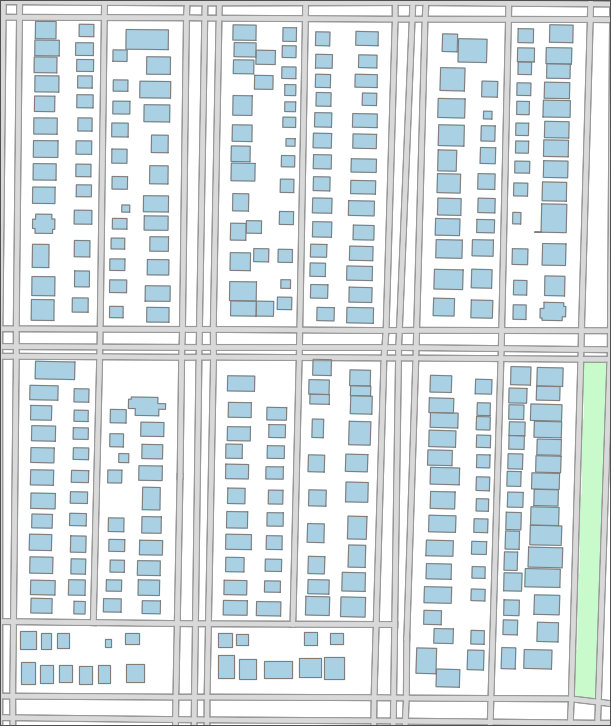} &
    \includegraphics[width=0.19\textwidth]{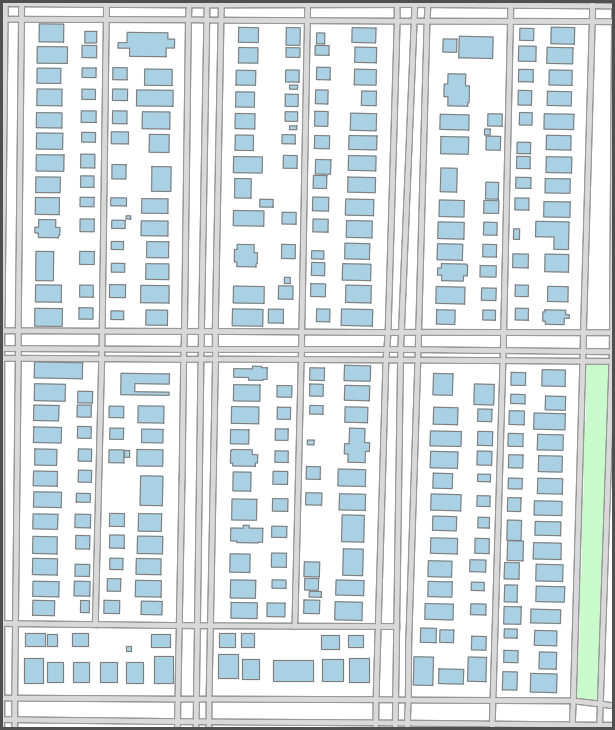} \\

    \includegraphics[width=0.19\textwidth]{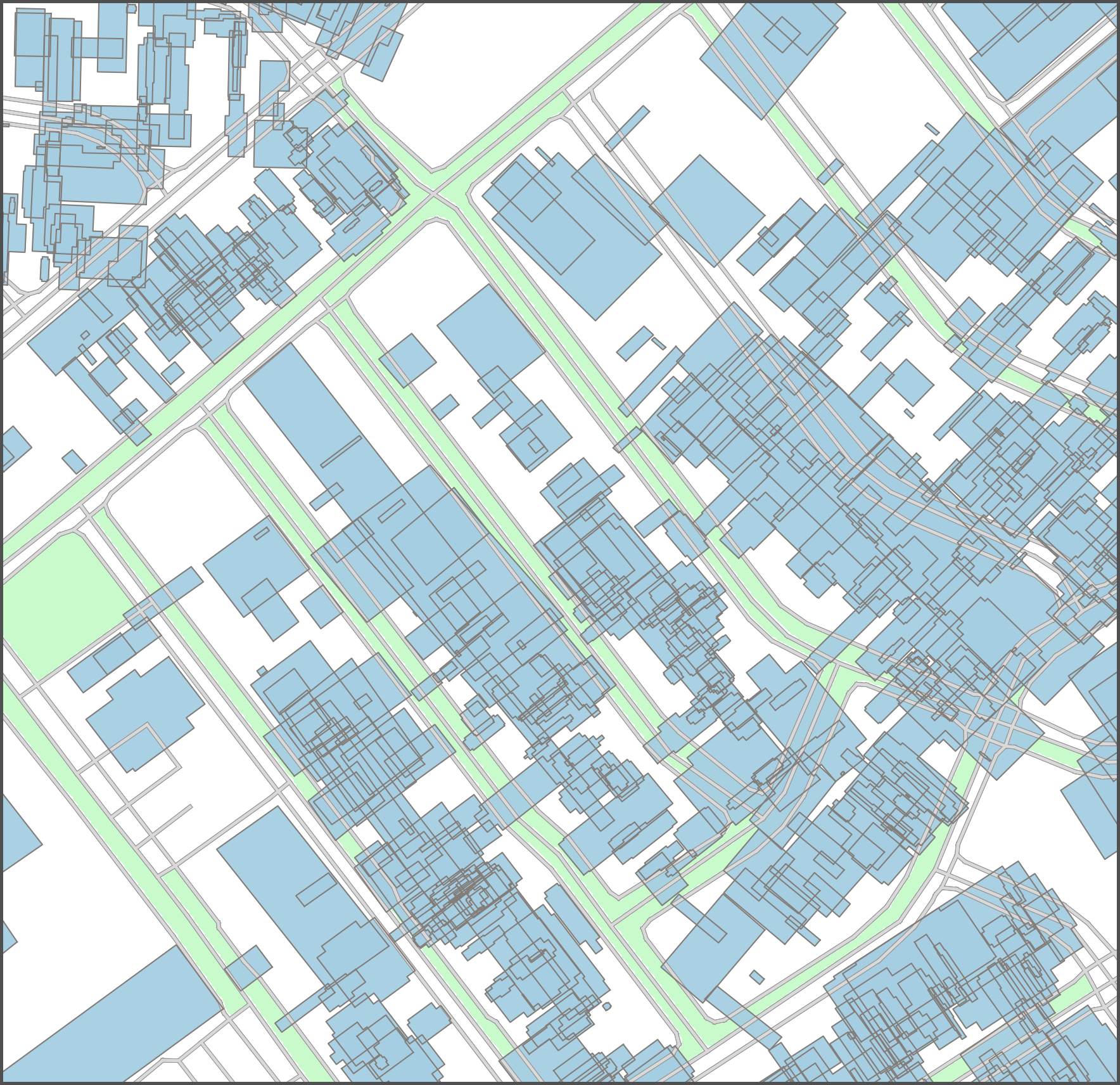} &
    \includegraphics[width=0.19\textwidth]{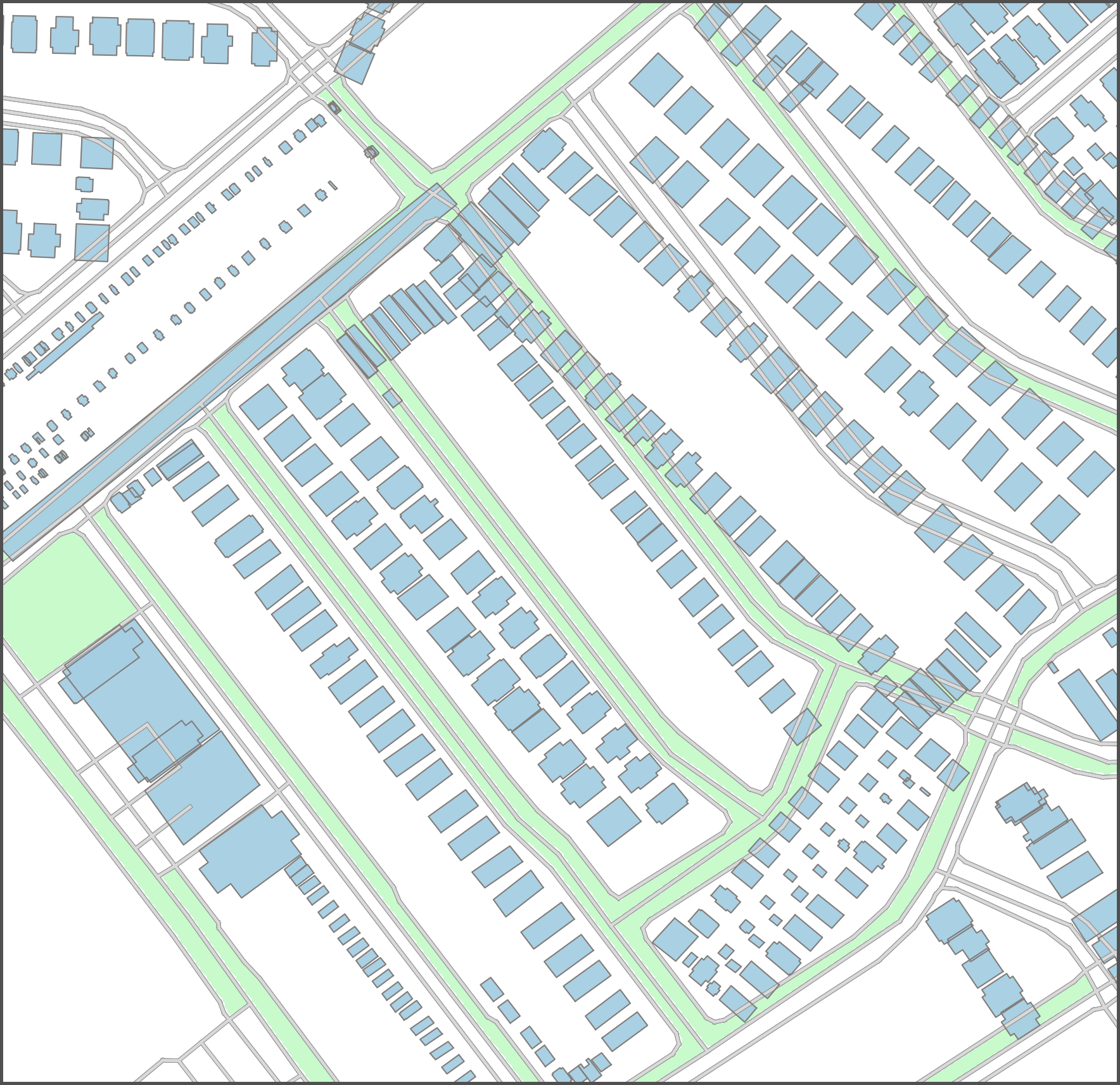} &
    \includegraphics[width=0.19\textwidth]{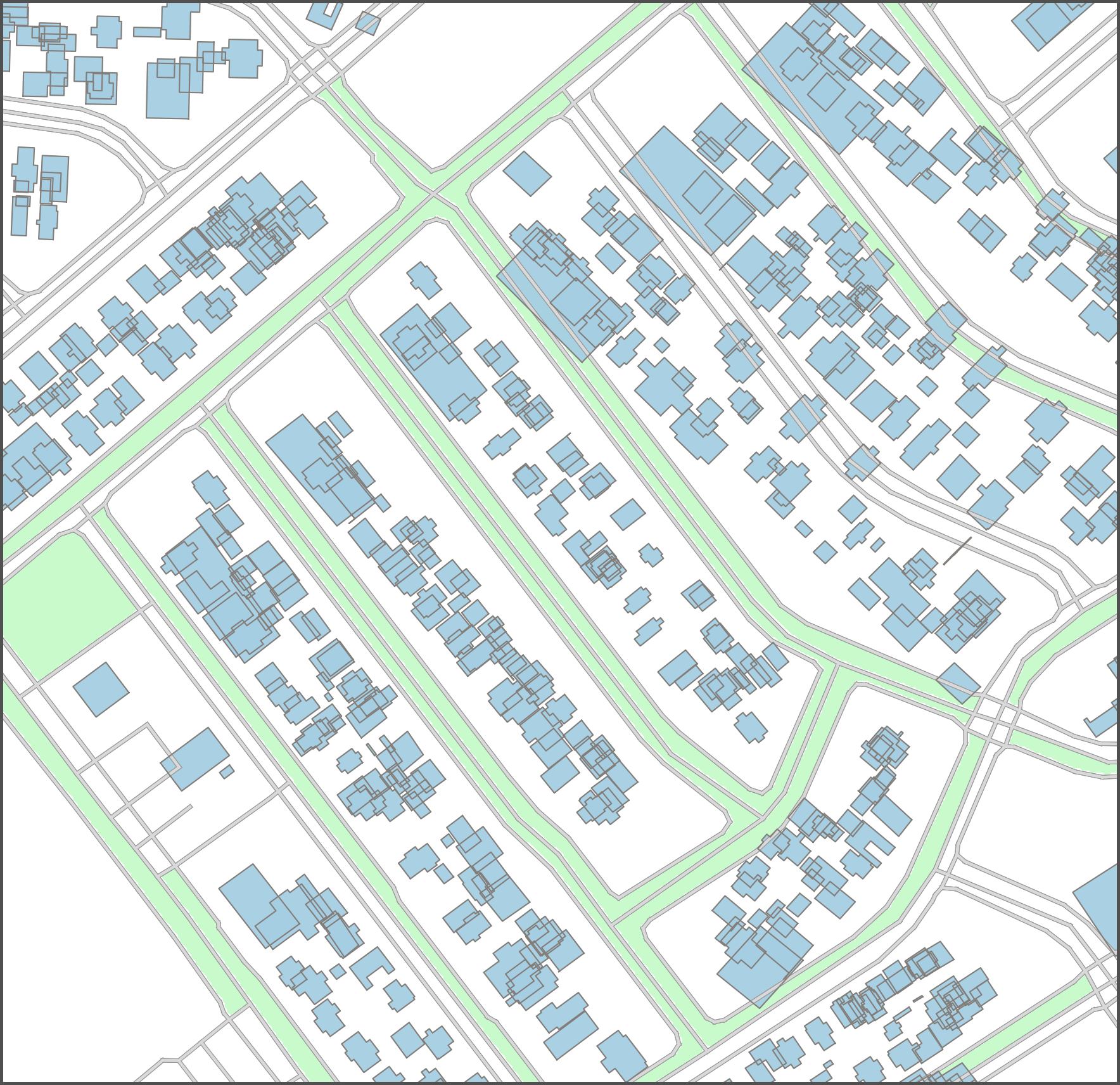} &
    \includegraphics[width=0.19\textwidth]{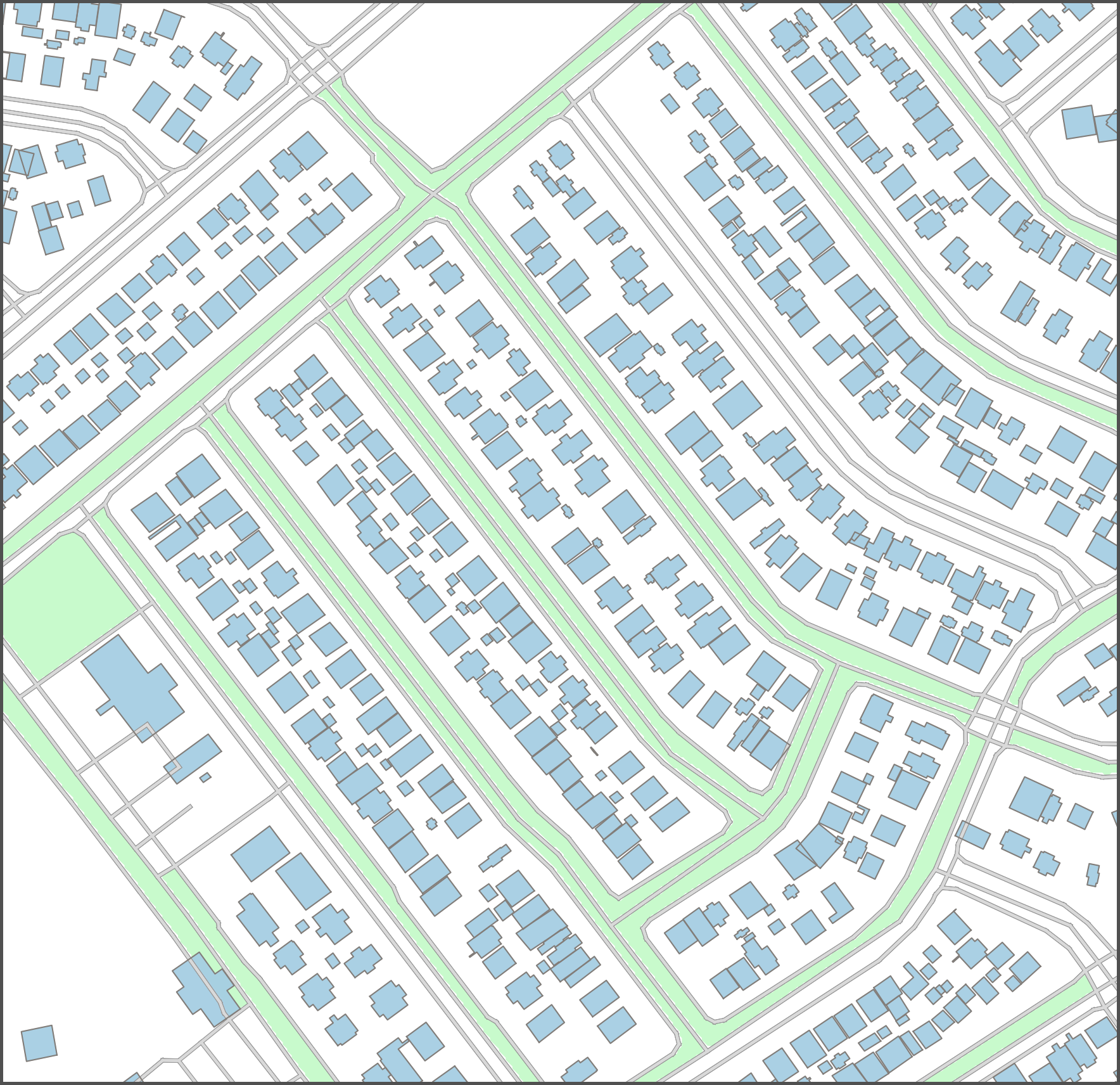} &
    \includegraphics[width=0.19\textwidth]{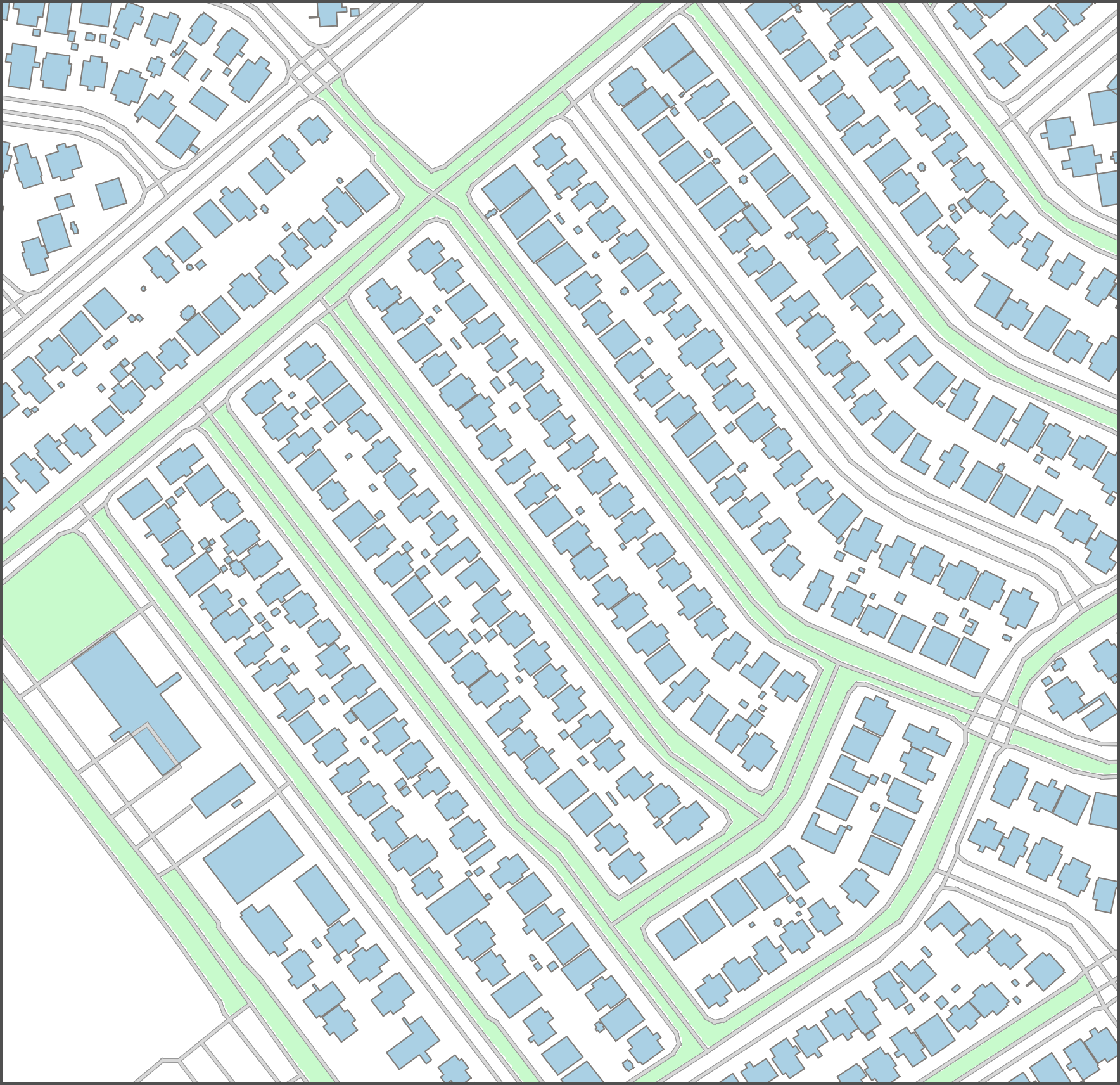} \\

    \includegraphics[width=0.19\textwidth]{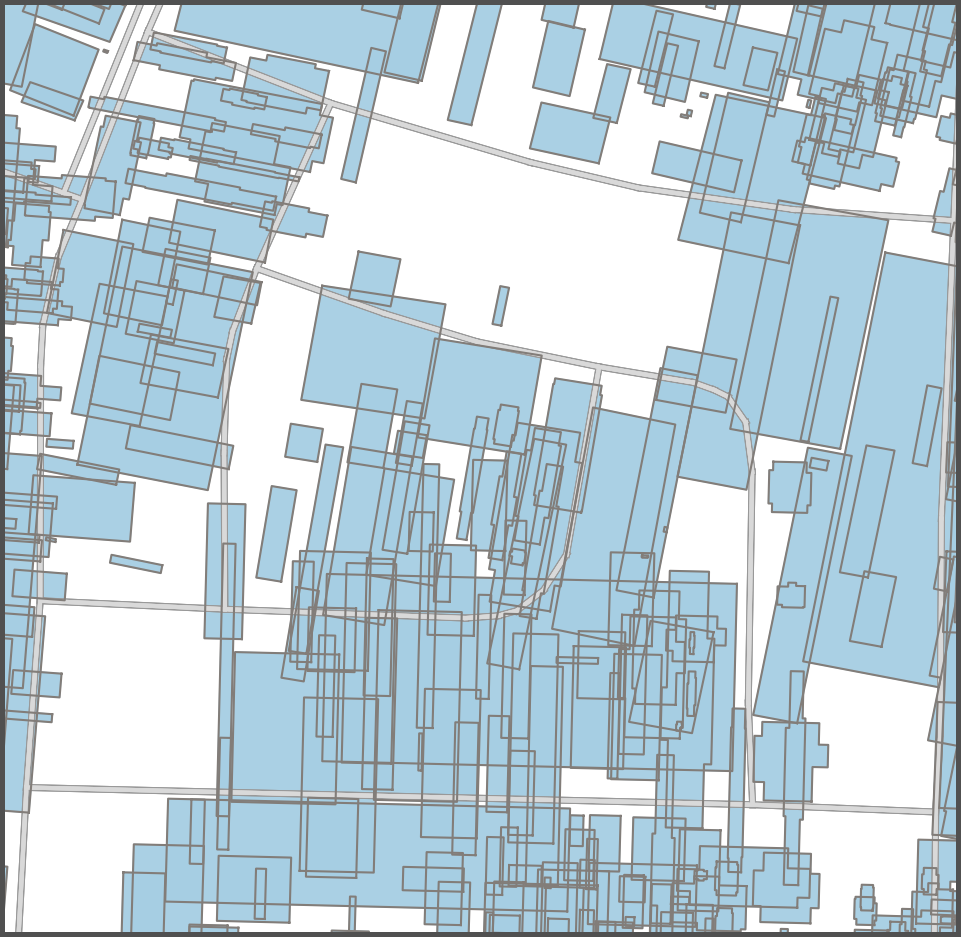} &
    \includegraphics[width=0.19\textwidth]{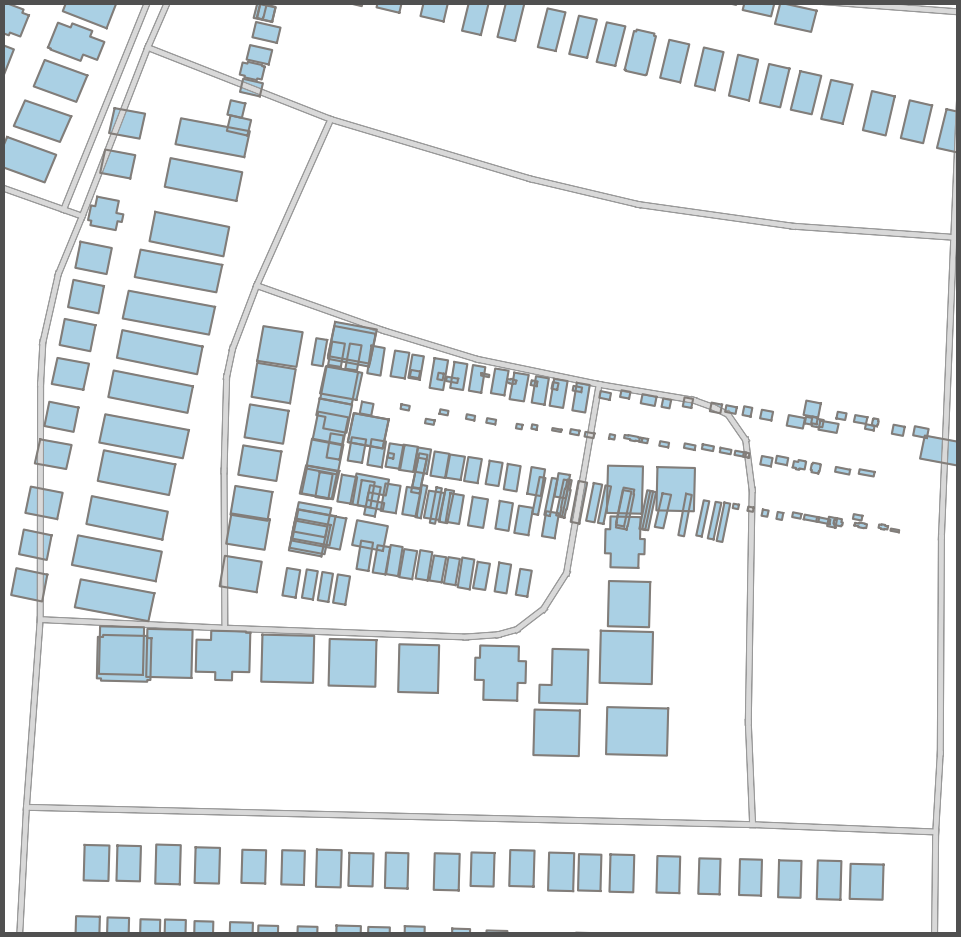} &
    \includegraphics[width=0.19\textwidth]{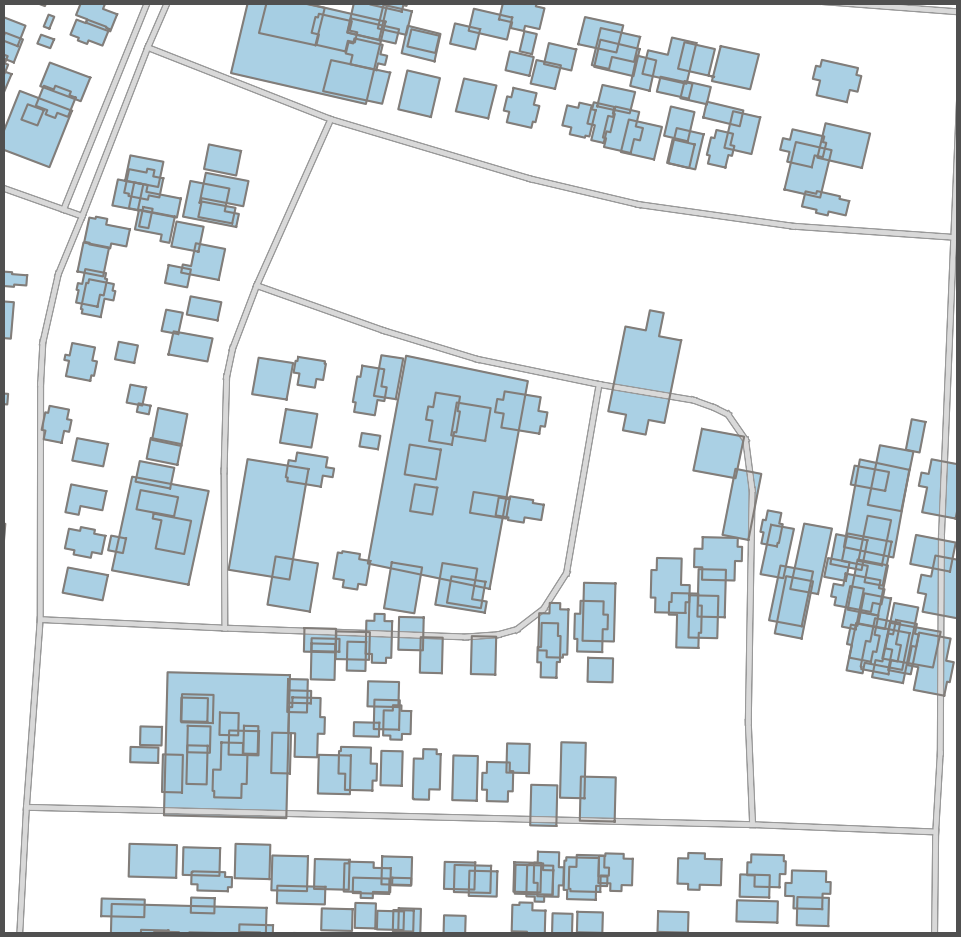} &
    \includegraphics[width=0.19\textwidth]{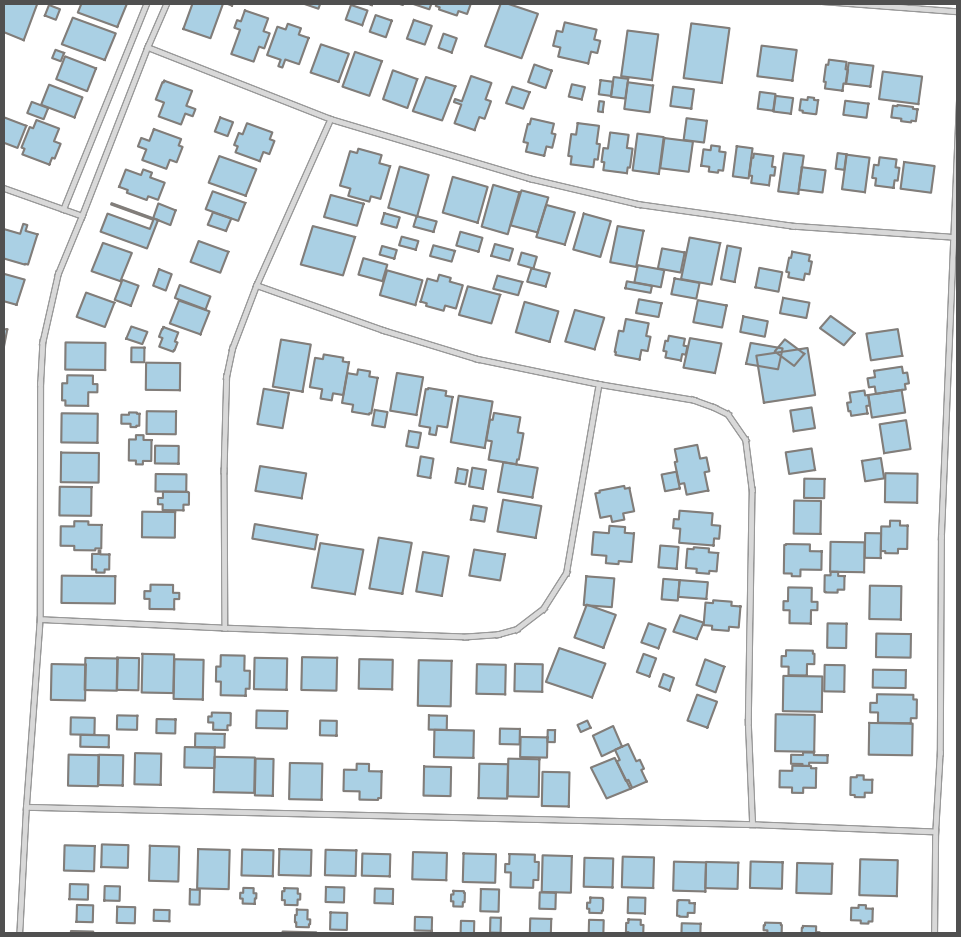} &
    \includegraphics[width=0.19\textwidth]{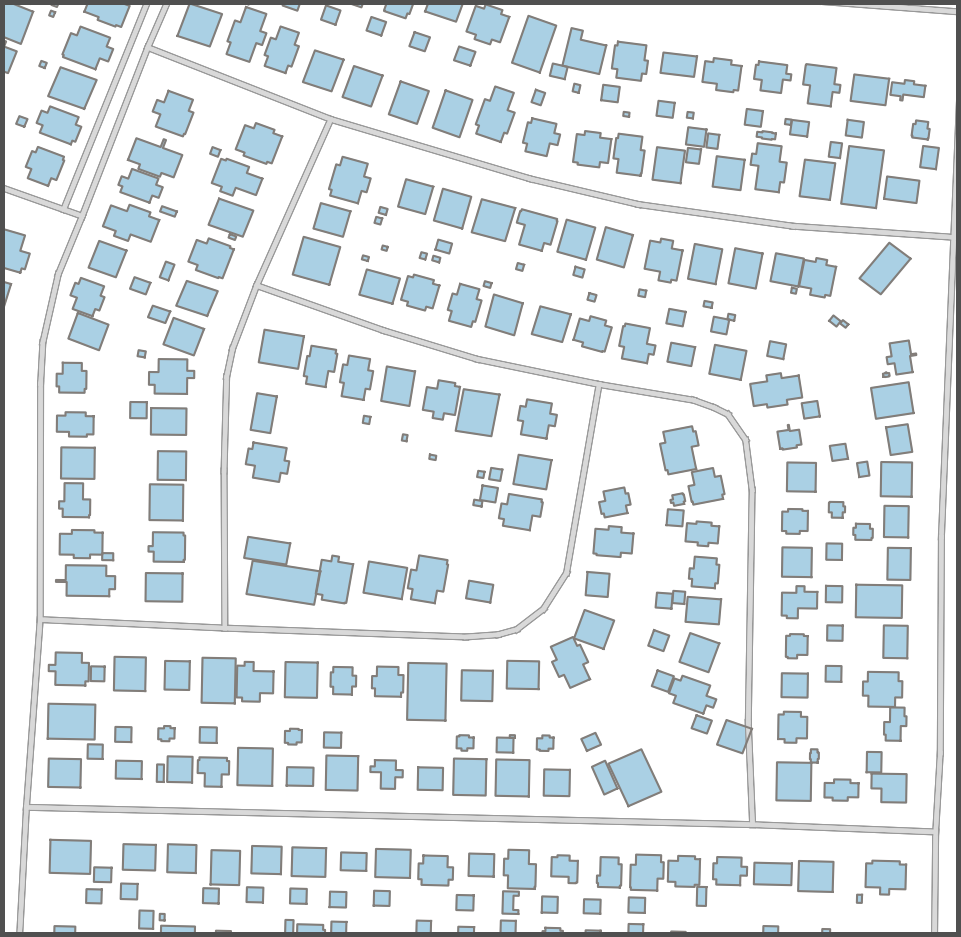} \\

    Layoutvae & VTN & BlockPlanner & Ours & Real Data
\end{tabular}%
  \vspace{-2mm}
    \caption{\textbf{Qualitative Results.} Given the same road network, all above methods generate urban layouts multiple times and we present the most similar one compared to the real data (similarity is evaluated by L-Sim index defined in Section~\ref{sec:comparison}. LayoutVAE~\cite{Jyothi19} fails in all types of block shapes. VTN~\cite{arroyo2021variational} produces rectilinear layouts regardless of block shapes. Blockplanner~\cite{xu2021} suffers overlapping and struggles under irregular block shapes. As compared to real data, our method is able to capture the layout styles more faithfully.}
  \vspace{-2mm}
  \label{fig:comparison}
\end{figure*}

\textbf{Qualitative Results.} Fig.~\ref{fig:comparison} shows urban layouts generated by our method and other methods (LayoutVAE~\cite{Jyothi19}, BlockPlanner~\cite{xu2021}, and VTN~\cite{arroyo2021variational}). For improved fairness, we provide ground truth building shape types and occupancy ratio values to our competitors since they did not consider building shapes. Nonetheless, the shown visual quality indicates that our method is still better able to capture arbitrary block shape and provide good generated output. In particular, Blockplanner~\cite{xu2021} struggles to produce densely distributed urban layouts as claimed. It might be due to two reasons. First, the model is retrained on a larger and diverse dataset (119,236 blocks, 28 cities) compared to the original paper (637 blocks, 1 city); second, our urban layout generation task differs to its original land lot generation task (i.e., most buildings are not supposed to touch each other as in their original case). VTN~\cite{arroyo2021variational} produces rectilinear building layouts regardless of block shapes. Strictly-rectilinear layouts are common in document layouts, but far from realistic in urban layouts.

\textbf{User Study.} We conducted a user study for perceptual realism using our method and the above three prior methods. The study was performed in a two-alternative forced choice (2AFC) manner. We generated 18 comparisons, each containing a layout generated by our method and the same layout generated by one of the three prior methods (thus 6 layout pairs for each prior method). The urban locations do not overlap across the three sets. During the study, we presented two urban layouts side-by-side from different methods. Users were asked to choose the one that looks more realistic to the best of their ability. We also included 2 random duplicate questions for quality checking. Users that answered differently to the same question were discarded. 

In total 62 survey results were received. We discarded replies that were finished too fast (less than 2 minutes), that always answered the same choice, and that did not pass the quality check. Replies from the remaining 50 valid users are summarized. 
Our method was preferred by 97.3\% of answers over LayoutVAE~\cite{Jyothi19}, 86.7\% over BlockPlanner~\cite{xu2021}, and 88.0\% over VTN~\cite{arroyo2021variational}.

\begin{table}[t]
\footnotesize
\centering
\begin{tabular}{lccccc}
    \toprule
   Ablations & Overlap$\downarrow$ & Out-Block$\downarrow$ & Pos-E.$\downarrow$ & Cov-E.$\downarrow$\\
    \midrule  
    LayoutVAE~\cite{Jyothi19} & 28.14 & 11.08 & 18.52 & 24.02 \\
    BlockPlanner~\cite{xu2021} & 5.87 & 3.47 & 8.04 & 8.41   \\
    Gupta \etal~\cite{gupta2021layouttransformer} & 3.54 & 7.43 & 10.50 & 0.49  \\
    VTN~\cite{arroyo2021variational} & 2.34 & 7.24 & 9.42 & 2.56  \\
    \midrule  
    \cite{Jyothi19} + CST & 28.00 & 8.12 & 15.09 & 24.34  \\
    \cite{xu2021} + CST & 3.20 & 2.08 & 4.09 & 6.11  \\
    \cite{gupta2021layouttransformer} + CST & 4.15 & 5.18 & 7.69 & 1.90   \\
    \cite{arroyo2021variational} + CST & 2.32 & 5.61 & 6.67 & 1.97   \\

\midrule  
    \cite{xu2021} + CST\&SE & 2.12 & 1.93 & 3.47 & 2.78  \\
    \cite{gupta2021layouttransformer} + CST\&SE & 3.89 & 4.34 & 5.65 & 1.73   \\
    \cite{arroyo2021variational} + CST\&SE & 2.25 & 4.52 & 5.89 & 1.88  \\

    \midrule  
   Ours - CST  & 4.86 & 2.23 & 4.40 & 3.94  \\
   Ours - SE     & 1.39 & 2.03 & 2.69 & 2.00 \\
   Ours - CST\&SE    & 7.09 & 3.03 & 7.87 & 9.88 \\

    \midrule  
   Ours (Xformer~\cite{shi2020masked})  & 1.08 & 4.40 & 6.00  &  0.38 \\
   Ours (SAGE~\cite{hamilton2017inductive})     & 1.36 & 5.04 &  6.84 &  0.73  \\
   Ours (GCN~\cite{kipf2016semi})      & 1.57 & 3.44 & 6.02 & 0.76 \\
   Ours (T=1)    & 1.38 & 1.30 & 3.41 &  0.88  \\
   Ours (T=2)     & 1.25 & 1.31 & 3.32  & 0.74  \\

    \midrule  
    \textbf{Ours* (T=3)} & \textbf{1.06} & \textbf{1.25} & \textbf{3.10} & \textbf{0.36} \\
    \bottomrule

\end{tabular}
    \vspace{-1mm}
  \caption{\textbf{Ablations.} We report metrics (all in $\%$) among all alternative ablations. $T$ is number of stacked graph attention layers. More ablations in Supplemental.}
    \vspace{-2mm}
  \label{tab:ablation}
\end{table}%


\subsection{Ablation Study} 
\label{sec: ablation}
We performed a comprehensive ablation study of adding and removing components of our solution to our method and to prior methods. While the previous comparisons show the superiority of our full method to prior methods as published, we further analyzed the effect of providing our canonical spatial transformation (CST) and also shape encoder (SE) to our competitors. In particular, CST was added to all competitors, and the vector produced by SE was also concatenated to the input vector of Gupta \etal~\cite{gupta2021layouttransformer} and to the latent bottleneck vector of VTN~\cite{arroyo2021variational} and BlockPlanner~\cite{xu2021} before decoding. For ablations on our model structure, we experimented with alternative graph convolutional layers (e.g. GraphSAGE~\cite{hamilton2017inductive}, transformer-based~\cite{shi2020masked}, and GCN~\cite{kipf2016semi}) with depth $T=3$. In addition, we varied attention depth ($T\in{1,2,3}$) in our model structure. 

We performed one-to-one comparisons using 1000 validation urban layouts, which is a comparison process different to the set-to-set comparison in Sec.~\ref{sec:comparison}. This more restrictive comparison task comprehensively evaluates robustness of overlap, position, and coverage rate. Position error is the average percent of building position shifting over the block length. Coverage error is the difference of total building area coverage rate compared to the ground truth rate.

Ablation results in Tab.~\ref{tab:ablation} illustrate that CST and SE both help robustness to arbitrary block shapes. It especially helps reduce Out-Block index and position error. This indicates our novelty has the potential to broadly benefit layout generation and improve adaptability to arbitrary canvas shapes. Moreover, even when both mechanisms are added to all competitors, our method still performs clearly better. Self-ablation results show that graph attention networks outperform other alternative convolutional layers, and the deeper structure improves model performance.




\subsection{Controllable Map Generation}
\label{sec:Control-Map}
Given a pre-trained model, we are able to generate diverse and realistic urban layouts for cities around the world. Apart from the random/conditional generation showed in Fig.~\ref{fig:teaser} and several more in Supplemental, we present several interesting generation options including sparse prior generation, semantic manipulation, and interpolation.

\begin{figure}[t]
\vspace{-1mm}
  \includegraphics[width=\linewidth]{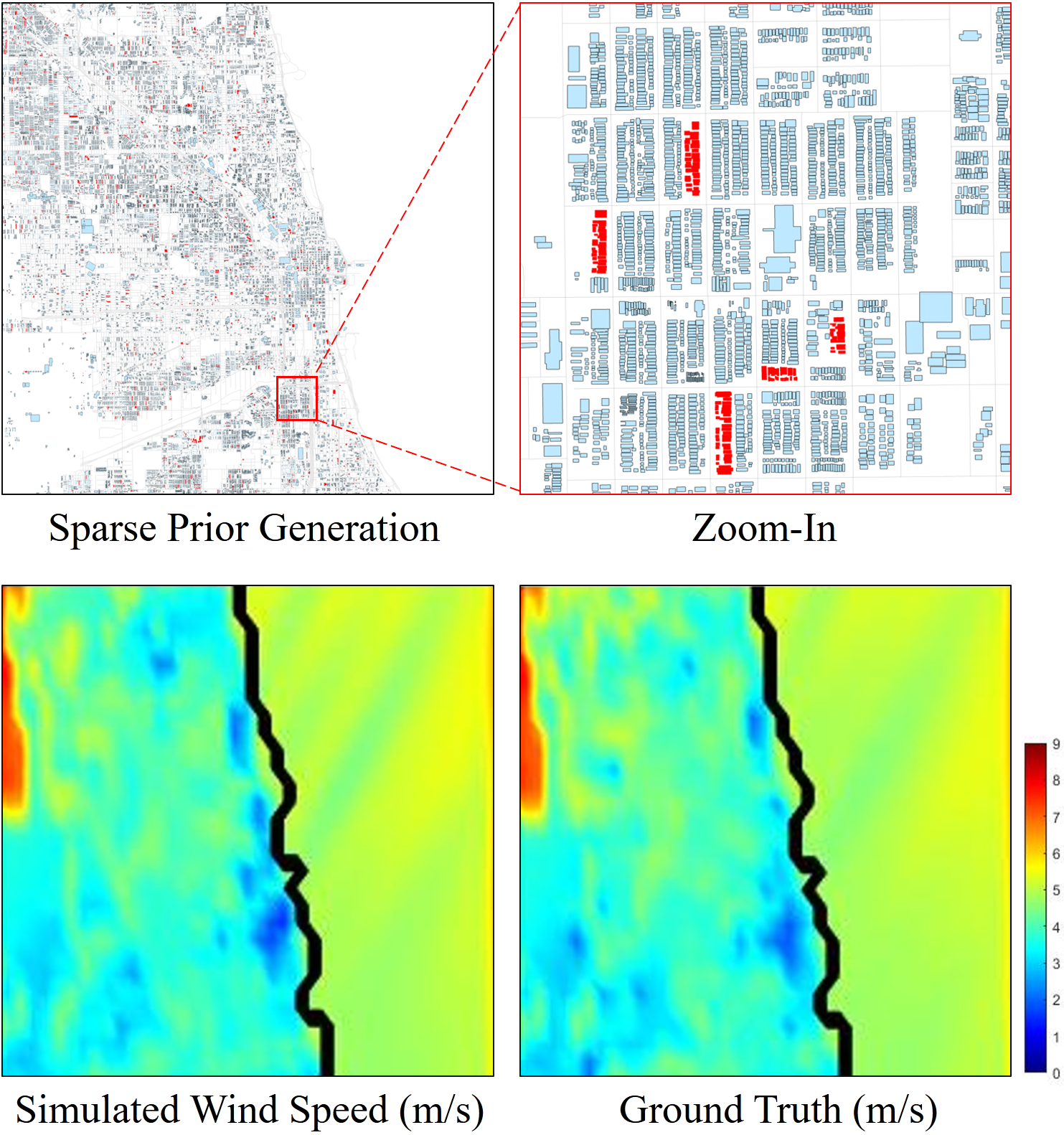}
  \caption{\textbf{Sparse Prior Generation.} Top: we show a generated Chicago from only 5\% prior data (shown in red), and a Zoom-In. Blank areas indicate non-building structures (e.g., parks, rivers, roads). Bottom: generated layout used by a local weather forecasting model~\cite{Chen2011} yields almost identical wind speed simulation compared to the ground truth.}
  \label{fig:sparse-prior}
    \vspace{-2mm}
\end{figure}

\textbf{Sparse Priors Generation.} Our approach supports conditioned generation given sparse building layouts as prior. In particular, given a road network and a few initial blocks of building layouts, we can generate more blocks with a similar style. Fig.~\ref{fig:sparse-prior} uses only a random 5\% of prior building layouts to generate a city: the latent vector of an empty block is calculated as the distance-weighted sum of the nearest $k$ blocks given as prior ($k=5$ works well). Normal noise is also added to the latent vector for diversity.

This generation ability is useful to provide data equity for many/most cities globally, where complete layout information is not available. To evaluate this, we showcased a well-modeled city. Fig.~\ref{fig:sparse-prior} models Chicago in which we trained building height information as additional building features so that we can generate 3D building masses (still from only 5\% of prior). Then, we extracted a 3D urban morphology suitable for an urban weather forecasting system, Urban WRF~\cite{Chen2011}. The model generated by our method from only 5\% prior yields comparable local weather forecasting results as that provided by the ground truth. Our average per-pixel wind-speed prediction error is 0.23m/s (details in Supplemental). Additional results are available in~\cite{patel2023deep}.


\textbf{Semantic Manipulation.} A semantic editing of building layouts can be performed by exploiting plausible disentanglements. For example, we labeled building row numbers (e.g., from 1 to 4) of 20K test urban layouts. We found latent vectors for each row-label group form reasonable clusters. If we translate a latent vector of a 1-row building layout towards that of the 2-row building layout cluster center, the layout will gradually have another row of buildings. Results are provided in Supplemental for up to 4 rows of buildings.


\textbf{Interpolation.} Our method can generate urban layouts by interpolating latent space vectors between two layouts. In Supplemental, we show results from linearly interpolating either building layout latent vectors or block shape latent vectors from one to another. The intermediate layouts correspond to an intuitive style interpolation.

\section{Conclusions and Future Work}
We have presented a controllable graph-based method to generate plausible urban layouts with arbitrary block shapes and varying building shapes. Our approach exploits multi-layer message passing using graph attention networks to encode and decode/generate the relationship between adjacent building structures. As opposed to image and pixel based methods, our technique exploits that urban building layouts are discrete, uses a stubby grid graph topology to support multiple rows of building structures, fits buildings to a taxonomy of parameterized building shapes in order to support a variety of building forms, and uses a skeletonization algorithm to map arbitrary city block shapes to a canonical form in order to better support a deep learning based approach. Our results, including user study, show superior performance to prior methods (e.g., LayoutVAE~\cite{Jyothi19}, BlockPlanner~\cite{xu2021}, Gupta \etal~\cite{gupta2021layouttransformer}, and VTN~\cite{arroyo2021variational}), analyzes different message passing schemes~\cite{kipf2016semi, hamilton17, prakash21, gat2018}, and demonstrates many examples in large cities. 

Our approach does have some limitations. First, our approach cannot represent city block contours with interior boundaries. Second, we do not support all possible building shapes. Third, only up to a fixed maximum number of buildings are supported in a block. Our method can handle common dead-end roads, such as cul-de-sac's (see Supplemental), but not all cases of dead-end roads.

As future work, we would like to ingest additional building semantics to produce more elaborate building structures, support further out-of-distribution layouts, and scale our method to cities worldwide.

{\small
\bibliographystyle{ieee_fullname}
\bibliography{egbib}
}

\end{document}